# Combining Deep Transfer Learning with Signal-image Encoding for Multi-Modal Mental Wellbeing Classification

Kieran Woodward[1,], Eiman Kanjo[1,], Athanasios Tsanas[1,]

**Abstract**

The quantification of emotional states is an important step to understanding wellbeing. Time series data from multiple modalities such as physiological and motion sensor data have proven to be integral for measuring and quantifying emotions. Monitoring emotional trajectories over long periods of time inherits some critical limitations in relation to the size of the training data. This shortcoming may hinder the development of reliable and accurate machine learning models. To address this problem, this paper proposes a framework to tackle the limitation in performing emotional state recognition on multiple multimodal datasets: 1) encoding multivariate time series data into coloured images; 2) leveraging pre-trained object recognition models to apply a Transfer Learning (TL) approach using the images from step 1; 3) utilising a 1D Convolutional Neural Network (CNN) to perform emotion classification from physiological data; 4) concatenating the pre-trained TL model with the 1D CNN. Furthermore, the possibility of performing TL to infer stress from physiological data is explored by initially training a 1D CNN using a large physical activity dataset and then applying the learned knowledge to the target dataset. We demonstrate that model performance when inferring real-world wellbeing rated on a 5-point Likert scale can be enhanced using our framework, resulting in up to 98.5% accuracy, outperforming a conventional CNN by 4.5%. Subject-independent models using the same approach resulted in an average of 72.3% accuracy (SD 0.038). The proposed CNN-TL-based methodology may overcome problems with small training datasets, thus improving on the performance of conventional deep learning methods.

*Email addresses:* `kieran.woodward@ntu.ac.uk` (Kieran Woodward), `eiman.kanjo@ntu.ac.uk` (Eiman Kanjo), `atsanas@ed.ac.uk` (Athanasios Tsanas)

# 1. Introduction

Monitoring and quantifying emotional states can potentially enable people to improve their wellbeing and self management as they understand their life stressors. Computational methods to infer emotional states based on physiological and environmental measurements require further exploration as with recent advances in wearable and sensor technologies along with machine learning algorithms, the real-time monitoring, collection and analysis of multi-model signals is becoming increasingly possible. Various ubiquitous sensors can be capitalized on to monitor physiological changes that affect wellbeing. The use of these sensors to measure diverse data modalities including Heart Rate (HR), Heart Rate Variability (HRV) and Electrodermal Activity (EDA) may enable real-world emotion recognition as they directly correlate to the sympathetic nervous system [1], [2].

Advances in Deep Learning (DL) present new opportunities for the inference of mental wellbeing as models can be trained using raw data, alleviating the need for manual feature extraction which is often domain-driven and may be a time-consuming process. Convolutional Neural Networks (CNNs) have traditionally been used to classify 2D data such as images but these networks are also employed towards extracting features from 1-dimensional sensor data [3]. However, the performance of machine learning models deteriorates considerably when training data is scarce, even more so with DL models. This lack of sufficient statistical power often hinders the progress of machine learning applications in monitoring and understanding wellbeing, since collecting longitudinal and annotated training data is very challenging [4]. This is due to the following reasons:

1. User availability, incentivisation and willingness to participate in longitudinal studies (or increasing study drop-outs beyond the first few months) [5]
2. Privacy, ethics and data protection issues [6], [7]
3. Data integrity and accuracy [8]
4. Costs and availability of monitoring devices [9], [10]
5. Requirement to set up the device and extract the data by expert personnel needing specialized equipment [11]
6. Time consuming nature of real-time self labelling [12]

In order to address these well reported problems, Transfer Learning (TL) is often used by training a base model using labelled data from a different domain and transferring the learned knowledge to the new target domain [13], [14]. Pre-trained models are often used to encompass methods that discover shared characteristics between prior tasks and a target task [14].

Previous research shows that fast changing, continuous sensor data such as accelerometer data can be transformed into RGB images which can then be used to train DL models [15]. Although the premise of presenting time series data as images is promising in extracting multi-level features and improving classification accuracy, most of the previous works only considered encoding univariate time series data as one image for a single channel of a CNN input [15], [16], [17],.

We aim to tackle the challenging problem of monitoring the trajectory of wellbeing by using DL and multiple datasets, including data we have collected from a controlled stressor experiment and real-world wellbeing experiment. Large wellbeing datasets are traditionally challenging to collect resulting in problems with the out-of-the-box application of existing tools. To address the limited sample size of wellbeing data we propose a new CNN-TL-based approach to alleviate many challenges when classifying small datasets. We explore the use of signal-image encoding to classify wellbeing using three techniques; Gramian Angular Summation Field (GASF), Gramian Angular Difference Field (GADF) and Markov Transition Field (MTF) [15]. We propose using these images in a novel pre-trained TL model combined with a 1D CNN trained using physiological sensor data to infer mental wellbeing. This framework uses TL in addition to signal-encoded images to improve model performance and generalise for new domains reducing the need for large datasets.

The remainder of the paper is organized as follows: Section 2 provides a review of mental wellbeing classification and TL; Section 3 describes and explores the three datasets used; Section 4 describes the data exploration, methodologies of data transformation and the model implementation; Section 5 shows the results, Section 6 presents the discussion and Section 7 presents the conclusion and suggestions for future research.

## 2. Related Work

### 2.1. Physiological Sensors to Monitor Wellbeing

Numerous non-invasive physiological sensors can be used to assess realworld mental wellbeing including EDA and HR [1]. EDA correlates to the sympathetic nervous system [2] and can be used to detect mental wellbeing while HRV is the variation in time between heartbeats: as HRV is reduced the user is more likely to be stressed [18]. Previous work had investigated the use of embedded sensors within a wearable device that measured EDA and HRV during driving [18]. The device took five minute recordings of physiological sensor data enabling the model to predict stress at 97.4% accuracy. HRV and EDA were found to be very well statistically associated demonstrating these non-invasive sensors have the capability to accurately infer mental wellbeing.

Previous research has also developed a wearable device that measured ElectroCardioGram (ECG), EDA and ElectroMyoGraphy (EMG) of the trapezius muscles [19]. 18 participants wore the device while completing three stressor tasks with a perceived stress scale questionnaire completed before and after each task. Stressed and non-stressed states were classified with an average accuracy of almost 80%. However, as this study was conducted in a controlled environment it is not known how well the model would generalize in real-world environments, where physiological signals may be impacted by more than just stress.

Skin temperature has also been explored to infer stress as it indicates acute stressor intensity [20]. A further study investigated a wearable device that measured EDA, skin temperature and motion. The devices were provided to six participants with dementia for two months with the ground truth labels obtained from clinical notes [21]. Stress was then assigned into one of five integer levels where accuracy ranged from 9.9% to 89.4% between the levels while F1-Scores ranged from 1.4% to 26.8% demonstrating a high level of false positives and false negatives. The wide variation of accuracy is due to the low stress threshold as when the threshold was increased there were fewer classifications of stress thus increasing accuracy.

*2.2. Mental WellBeing Classification using Deep Learning*

Once physiological data has been collected from the devices, different models must be explored to assign class labels. There are two main types of neural networks: Convolutional Neural Networks (CNNs) and Recurrent Neural Networks (RNNs). They are structurally different and are used for fundamentally different purposes. CNNs have convolutional layers to transform data, whilst RNNs reuse activation functions from other data points. Previously, RNN Long Short-Term Memory (LSTM) networks have been used to classify mental wellbeing as they capture long-term temporal dependencies. An LSTM network with a stack autoencoder to decompose the combined EEG signals was used to infer emotions from 32 participants. This approach of using the context correlations of the EEG feature sequences resulted in increased performance, achieving 81.1% accuracy [22]. Furthermore, previous work has fused raw EEG signals with videos of participants to improve model accuracy when inferring wellbeing [23]. The model achieved 74.5% accuracy by using temporal attention to ignore the redundant information. LSTM networks have also been used to classify EDA, skin temperature, motion and phone usage data to infer stress achieving 81.4% accuracy, outperforming other support vector machine and logistic regression models
[24].

CNNs have also been used to infer mental wellbeing. A CNN has been trained to classify four emotions (relaxation, anxiety, excitement and fun) using EDA and blood volume pulse data [25]. DL algorithms were compared with standard feature extraction and selection approaches concluding DL outperformed manual ad-hoc feature extraction as it produced significantly more accurate affective models, even outperforming models that were boosted by automatic feature selection. Additionally, a CNN model using channel selection strategy has been trained using EEG data collected from 32 participants watching 40 1-minute excerpts of music videos to elicit emotions [26]. The channel selection strategy used the channels with the strongest correlation with valence to generate the training set. The model classified four possible emotions: (1) high arousal and high valence (2) high arousal and low valence (3) low arousal and high valence (4) low arousal and low valence. Using the EEG data, this channel selection approach achieved 87.27% accuracy, improving the accuracy by nearly 20%. A CNN and an RNN have been combined to allow raw data to be classified more accurately automating feature extraction and selection [27] [28]. Physiological, environmental and location data was used to train the model to infer emotions resulting in the combined model outperforming traditional DL models by over 20%. This work concluded that the CNN model matched or outperformed models with the features pre-extracted showing the benefits of DL.

## 2.3. Transfer Learning

One of the biggest challenges in developing accurate DL models is the implicit practical requirement to collect a large *labelled* dataset. TL [14] is a common approach in machine learning to mitigate the problem occurring due to the scarcity of data. Caruana [13] introduced multi-task learning that uses domain information contained in the training signals of related tasks. It is based on the ability to learn new tasks relatively fast, alleviating the need for large datasets by relying on previous, similar data from related problems. TL capitalizes on a likely large dataset stemming from a related problem to pre-train a model, and subsequently adapt that model for the needs of a problem with a (potentially smaller) different dataset [29]. CNNs are commonly used in TL approaches, being initially trained on a vast dataset and then having the last fully-connected layer removed and further trained on a smaller target dataset. A pre-trained CNN alleviates the need for a large dataset while simultaneously decreasing the time required to train the model. The premise of TL is to improve the learning of a target task in three ways [30]: (1) improving initial performance, (2) producing sharp performance growth, (3) potentially resulting in higher training performance.

Hitherto, TL has most commonly been used to train images as large ImageNets have been used to developed pre-trained models such as VGGNet [31], Inceptionv3 [32] and mobileNetv3 [33] that contain pre-trained object classification models. The pre-trained CNN models were employed to compute mid-level image representations for object classification in PASCAL VOC images [34], leading to significantly improved results. TL has facilitated training new models in the visual domain using pre-trained CNNs [35]. However, modelling emotions using time series data such as HRV, HR, EDA or acceleration cannot be visually interpreted. Sensor data must first be transformed to translate the raw sensor data to images, for example using techniques such as GASF, GASF and MTF.

TL approaches have previously been applied to emotion assessment. One novel approach used a sparse autoencoder-based feature TL approach to infer emotions from speech using the FAU Aibo Emotion Corpus dataset containing 6601 instances of positive valence and 3358 instances of negative valence in the training set. The autoencoder approach was used to find a common structure in the small target base dataset and apply the structure to source data, improving average recall from 51.6% to 59.9% using only 50 data instances [36]. Whispered speech has also been explored to infer emotions applying three TL approaches; denoising autoencoders, shared-hidden-layer autoencoders, and extreme learning machines autoencoders. The extreme learning machines autoencoders demonstrated the best performance enhancing the prediction accuracy on a range of emotion tasks, achieving up to 74.6% arousal [37]. Speech has also been explored to improve post-traumatic stress disorder diagnosis using TL and deep belief networks. The TL approach transferring knowledge learned from a large speech database improving model accuracy from 61.53 to 74.99% [38].

TL for emotion recognition has also been used to assess mental health status from messages posted on Twitter [39]. A RNN with full weight transfer using a dataset to classify tweets achieved an overall accuracy of 78% for four classes, a 6% improvement over a standard RNN. Inter-subject TL approaches have been used with time-series data such as ECG signals, achieving 79.26% compared with a baseline of 67.90%, again demonstrating the potential of TL to improve model performance with small datasets [40].

The inference of emotions from images and videos has also benefited from TL approaches. When using pre-trained ImageNet models and transferring this knowledge to infer seven facial expression an accuracy of 55.6% was achieved compared with the baseline performance of 39.13% [41]. Additionally, audio and video have been explored to infer six emotions where the TL approach improved base line accuracy by 16.73% [42].

## 3. Data

In this work, three different datasets have been used containing data from physiological, environmental and motion sensors, along with their labels. The first dataset includes physiological data from a controlled stressor experiment, the second collected physiological and environmental real-world data along with self-reported labels and the third is a publicly available human activity dataset containing accelerometer data. The details of these three datasets are described in the following sections.

### 3.1. Dataset 1: Controlled Stressor Experiment

Experimental setup: We conducted a lab-based stressor experiment where the participants' various emotions were stimulated using the Montreal stress test [43]. This experiment induced stress in 20 healthy participants aged 18-50 between June - September 2019 as approved by Nottingham Trent University human ethics board, application number 600. To allow the effects of stress and mental arithmetic to be investigated separately the experiment has 3 test conditions; rest, control and experimental. Each participant was initially briefed before completing a 3-minute rest period where participants looked at a static computer screen where no tasks were displayed. This was followed by 3 minutes of the control condition where a series of mental arithmetic tasks were displayed which participants answered, followed by another 3-minute rest period. Participants then completed the stressor experiment where the difficulty and time limit of the tasks was adjusted to be just beyond the individual's mental capacity. The time pressure along with a progress bar showing their progress compared with an artificially inflated average were both designed to induce stress during the 10-minute experiment. Finally, participants completed a final 3-minute rest session and answered questions on their subjective experience of task load (NASA Task Load Index) [44], mental effort (Rating Scale Mental Effort) [45], emotion (Self-Assessment Manikin) [46] and stress (visual analogue scale) [47].

Sensors: Participants wore hand-held non-invasive sensors on their fingers, which recorded HR, HRV, and EDA each sampled at 30Hz to collect physiological data while experiencing relaxed and stressed states of mental wellbeing.

*3.2. Dataset 2: EnvBodySens*

Experiment setup: We previously collected the EnvBodySens dataset [25], which consists of 26 data files collected from 26 female participants (average age of 28) walking around the city centre in Nottingham, UK on specific routes. The participants were asked to spend no more than 45 minutes walking in the city center. Data was collected in similar weather conditions (average 20∘C), at around 11am. Participants were asked to periodically report how they feel based on a 5-point predefined emotion scale as they walked around the city centre using a smartphone app developed for the study. We adopted the 5-step SAM Scale for Valence taken from Banzhaf et al. [48] to simplify the continuous labelling process with each participant entering an average of 134 self-reports. During the data collection process, 550,432 sensor data frames were collected as well as 5345 self-report responses rated from 1 (most positive) to 5 (most negative), where sufficient samples for each rating were collected (1-8.44%, 2-9.95%, 3-29.91%, 4-21.71%, 5-29.99%). We disabled the screen auto sleep mode on the mobile devices, so the screen was kept on during the data collection process [25]. Data from six users were excluded due to logging problems. For example, one user was unable to collect data due to battery problem with the mobile phone, another user switched the application off accidentally.

Sensors: The dataset is composed of non-invasive physiological data (HR, EDA, body temperature, acceleration) sampled at 8Hz, environmental data (noise levels, UV and air pressure) also sampled at 8Hz, time stamps and self reports. The data was logged by the EnvBodySens mobile application on Android phones (Nexus), connected wirelessly to a Microsoft wrist Band 2 [49] that was provided to participants to collect the physiological and environmental data.

*3.3. Dataset 3: The Human Activity Recognition Using Smartphones Data Set (WISDM)*

Experiment setup: We used the dataset previously published by Kwapisz JR, Weiss GM, Moore SA (2011) [4]. The dataset comprises of 36 participants performing six physical activities (walking, walking upstairs, walking downstairs, sitting, standing and jogging) while carrying the provided Android-based smartphone in their front trouser leg pocket. The 36 participants resulted in 1,098,207 data samples collected consisting of 39% walking, 31% jogging, 11% walking upstairs, 9% walking downstairs 6% sitting and 4% standing.

Sensors: The dataset comprises of accelerometer data from the embedded sensor within the Android phone sampled at 20Hz. A graphical user interface was included within the custom Android application enabling users to label the 6 prespecified physical activities (walking, jogging, walking upstairs, walking downstairs, sitting and standing) when commencing a new activity and to end the data collection.

This external dataset is freely available at: http://www.cis.fordham.edu/wisdm/dataset.php

## 4. Methods

Using the three aforementioned datasets, we propose two TL approaches to help improve accuracy and reduce the requirement of large datasets when classifying mental wellbeing:

1. Training a base model on a large human activity dataset and transferring the learned knowledge to a 1D CNN trained using the controlled experiment dataset to infer stress.
2. Using signal-image encoding to transform accelerometer data into images and then applying a novel CNN-TL-based approach combined with a separate 1D CNN trained using the remaining sensor data from the EnvBodySens dataset.

*4.1. Data Exploration*

Controlled Stressor Experiment: The controlled experiment dataset resulted in a total of 417251 sensor data frames when relaxed and 475232 data frames when stressed. Figure 1 shows a sample of 400 samples representing around 13 seconds of relaxed and stressed data during the experiment. The dataset contains HR (mean 79.4BPM, SD 11.6BPM), HRV (mean 773.8ms, SD 152.6ms) and EDA (mean 320.4kΩ, SD 158kΩ) physiological sensor data but in comparison with the EnvBodySens dataset (see Dataset 2 used in this study) lacks motion data as participants were still while completing the experiment. When participants were resting the average HR was 76.9 (SD 10.5) compared with 81.6 when stressed (SD 12.1) showing an elevated HR when stressed. Similarly HRV reduced from 795.4ms (SD 149ms) when resting to 754.9ms (SD 152.6ms) when stressed. Finally EDA significantly reduced when stressed from 345.3kΩ (SD 152.6kΩ) to 298.6kΩ (SD 159.5kΩ) showing similar trends to the EnvBodySens dataset.

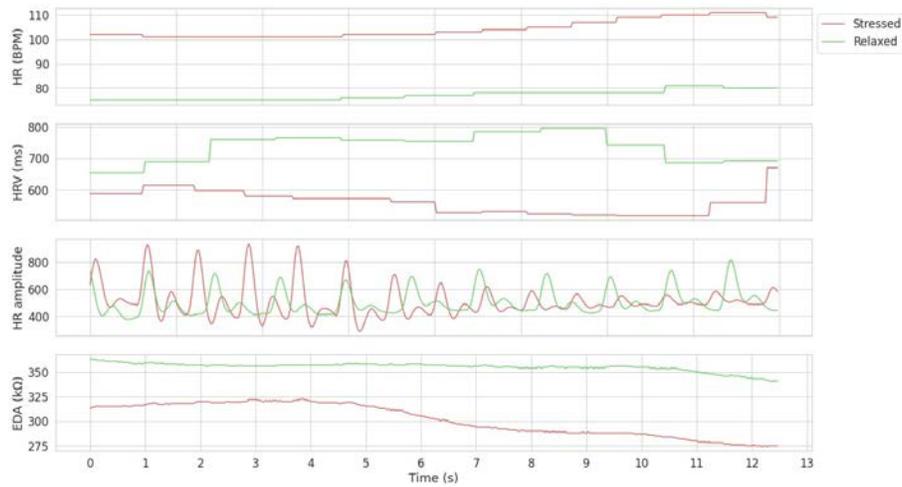

Figure 1: Comparison of stressed and relaxed data collected from the controlled stressor experiment.

Figure 2 shows that when stressed the distribution of the HR and EDA data is highly concentrated within two clustered areas compared with the relaxed data which is more sporadically dispersed. The dispersed data demonstrates that HR may still be high and EDA low when relaxed but it is not as concentrated as when stressed. Overall, the trends from the dataset show that EDA is an important indicator of stress and when paired with HR and HRV can show clear patterns of stressed and relaxed emotions.

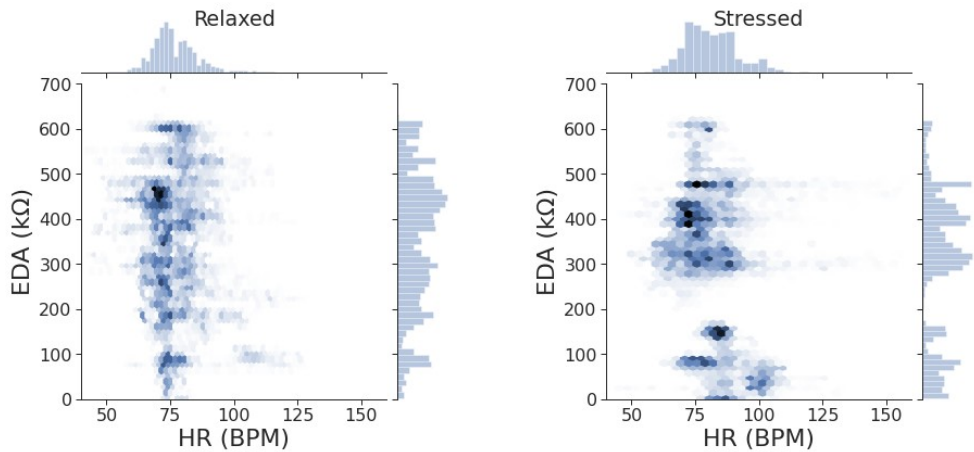

Figure 2: Summarizing the controlled experiment data when participants were relaxed (left panel) vs data when participants were stressed (right panel).

EnvBodySens: The EnvBodySens dataset contains only HR and EDA physiological data. The dataset resulted in 29965 data frames for state 1, 35333 frames for state 2, 106210 frames for state 3, 77103 frames for state 4 and 106478 frames for state 5. Figure 3 shows the EDA (mean 1455.2kΩ, SD 2870.5kΩ) and HR (mean 74.5BPM, SD 11.8BPM) for all participants when experiencing each of the five self-reported states of valence from 1 being most positive to 5 being most negative.

The distribution in figure 3 demonstrates that as users record higher levels of stress the average EDA value decreases. The EDA data collected behaves as expected with the median EDA value gradually decreasing as users become more stressed. However, stress levels do not impact the distribution of heart rate like EDA, instead the distribution of HR remains relatively similar for all stress levels. Stress level 2 has the highest distribution of HR with the high distribution reaching over 120 Beats Per Minute (BPM) even though this is the second most relaxed state. As users become more stressed the upper adjacent values are reduced all of which is unexpected as when users are stressed they are more likely to have increased heart rate [50]. The outlier HR data in states 1 and 2 that go beyond 180BPM are most likely artifacts of the data due to sensor error demonstrating that there is little change in HR over the 5 states of wellbeing. Overall, the EDA data behaves as expected while there is little to distinguish HR during the different states of wellbeing.

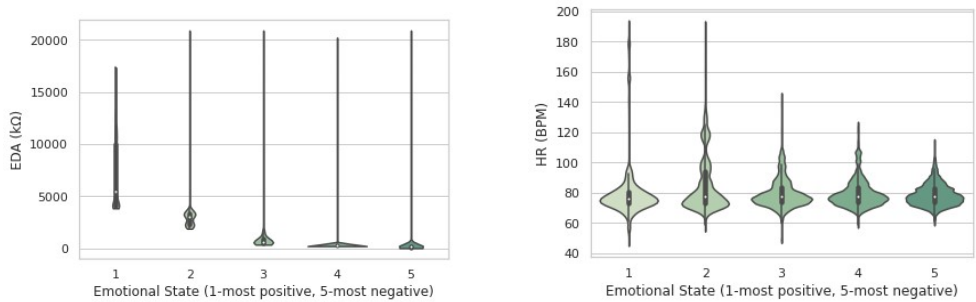

Figure 3: EnvBodySens EDA and HR data for reported emotional states from 1 (positive) to 5 (negative).

*4.2. Modality Transformation*

An image is comprised of pixels which can be conveniently represented in a matrix with a colour image containing three channels; red, green and blue for each pixel, compared with grayscale images that contain only one channel.

Transforming time series data into images helps extract multi-level features [15] and improve classification accuracy [16], [17].

This study aims to explore the use of 2D CNNs and the addition of TL with time series data. Therefore the continuous, fast changing datastream of accelerometer data must first be transformed into images. It is not plausible to transform physiological data into images due to its static nature where HR and EDA can often remain constant for several seconds resulting in no data being encoded. Three methods of modality transformation using accelerometer data are utilised: GADF, GASF and MTF.

Wang and Oates transformed time series data into images using Gramian Angular Field (GAF) [15]. First, the data was normalised between -1 and 1 by applying (1). The normalised data is then encoded using the value as the angular cosine and the time stamp as the radius $r$ with (2), where $\phi$ is the angle polar coordinates; $ti$ is the time stamp, $N$ is a constant factor to regularize the span of the polar coordinate system and $\tilde{X}$ represents the rescaled time series data [16].

$$\tilde{x}^i_{-1} = \frac{(x_i - \max(x)) + (x_i - \min(x))}{\max(x) - \min(x)} \quad (1)$$

$$\begin{cases} \phi = \arccos(\tilde{x}_i), -1 \leq \tilde{x}_i \leq 1, \tilde{x}_i \in \tilde{X} \\ r = \frac{t_i}{N}, t_i \in \mathbb{N} \end{cases} \quad (2)$$

$$GASF = [\cos(\emptyset_i + \emptyset_j)] \quad (3)$$

$$= \tilde{X}' \cdot \tilde{X} - \sqrt{I - \tilde{X}^2}' \cdot \sqrt{I - \tilde{X}^2} \quad (4)$$

$$GADF = [\sin(\emptyset_i - \emptyset_j)] \quad (5)$$

$$= \sqrt{I - \tilde{X}^2}' \cdot \tilde{X} - \tilde{X}' \cdot \sqrt{I - \tilde{X}^2} \quad (6)$$

The normalized data is then transformed into polar coordinates instead of the typical Cartesian coordinates. After transformation, the vectors are transformed into a symmetric matrix called the Gramian Matrix. There are two ways to transform the vectors into a symmetric matrix: GASF and GADF as shown from (3) to (6) where $\emptyset$ is the angle polar coordinates. These methods preserve the temporal dependency, with the position moving from top-left to bottom-right with time.

$$M = \begin{bmatrix} w_{ij}|x_1 \in q_i, x_1 \in q_j & \cdots & w_{ij}|x_1 \in q_i, x_n \in q_j \\ w_{ij}|x_2 \in q_i, x_1 \in q_j & \cdots & w_{ij}|x_2 \in q_i, x_n \in q_j \\ \vdots & \ddots & \vdots \\ w_{ij}|x_n \in q_i, x_1 \in q_j & \cdots & w_{ij}|x_n \in q_i, x_n \in q_j \end{bmatrix} \quad (7)$$

Alternatively, images can be generated using MTF where the Markov matrix is built and the dynamic transition probability is encoded in a quasiGramian matrix as defined in (7). Given a time series $x$ and its $q$ quantile bins each $x_i$ is assigned to the corresponding bins $q_j$ ($j \in [1,q]$). A $q$ X $q$ Markov transition matrix ($w$) is created by dividing the data into $q$ quantile bins. The quantile bins that contain the data at time stamp $i$ and $j$ (temporal axis) are $q_i$ and $q_j$. The information of the inter-relationship is preserved by extracting the Markov transition probabilities to encode dynamic transitional fields in a sequence of actions [15]. A comparison of identical X, Y, Z and average accelerometer data transformed as GASF, GADF and MTF can be seen in figure 4.

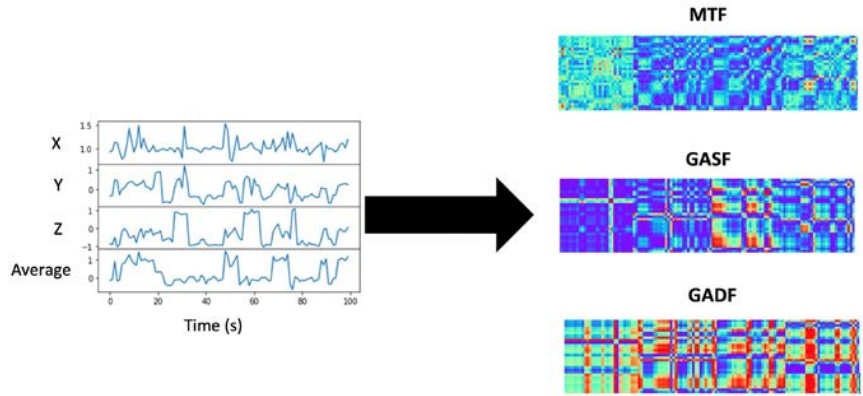

Figure 4: An example of raw accelerometer data (X, Y, Z and average motion) transformed using MTF, GASF and GADF.

*4.3. Computational Models*

We explore two TL approaches to classify the mental wellbeing data collected from the controlled stressor experiment and the EnvBodySens datasets. We used TensorFlow to implement our models on 2 Tesla M60 Graphical Processor Units (GPU) (16 GB memory, 899 MHz base clock speed, 4096 cores per GPU).

*4.3.1. 1-Dimensional CNN TL Model*

CNNs are feed forward networks that like RNNs do not require features to be first extracted although unlike RNNs they extract local and positional invariant features rather than different patterns across time. 1D CNNs utilise a feed-forward structure like 2D CNNs, although 1D CNNs process 1-dimensional patterns with 1-dimensional convolution operations, while 2D CNNs processes 2-dimensional patterns with 2-dimensional convolutions.

We first train a 1D CNN using the accelerometer data from the WISDM dataset for activity recognition and then apply a TL approach to transfer the learned knowledge towards the smaller controlled stressor dataset where the model is re-trained using the HR, HRV and EDA data to classify stressed and non-stressed states. This approach should help reduce overfitting and improve the accuracy in which stress can be classified from a limited sample by transferring the weights from the base model.

The proposed 1D CNN has three layers; an input layer, an output layer and a hidden layer. The training input data is represented as $x = [x_1, x_2, x_j]$, where the number of training samples is j and y is the output vector [51]. When $\sigma$ is the sigmoid activation function, w1 and w2 are weight matrices between the input and hidden layer and the hidden and output layer respectively. Finally, $b_1$ and $b_2$ represent the bias vectors of the hidden and output layer respectively [52]:

$$h = \sigma(w_1 x + b_1] \quad (8)$$

$$y = \sigma(w_2 h + b_2) \quad (9)$$

Batch normalisation has been used within the network to normalise the inputs of each layer so they have a mean of 0 and standard deviation of 1 this enables the models to train quicker, allows for higher learning rates and makes the weights easier to initialise [53]. An overlapping sliding window strategy has been adopted to segment the time series data with a window size of 100 and a step of 20 chosen experimentally, by trying different window sized from 10 to 400. A dropout layer with a rate of 0.5 was added before the maxpooling layer to prevent overfitting by randomly ignoring selected neurons during training [54]. The pooling layers subsample the data, reducing the number of weights within that activation. Finally, the fully-connected layers where each neuron is connected to all the neurons in the previous layer are used to calculate class predictions from the activation. Hold-out validation is used to test this approach by using a randomly selected 20% test split (around 178,496 data frames) to test the model and calculate the F1-Score, a harmonic average of the precision and recall.

*4.3.2. Image Encoding TL Model*

We propose the novel combination of a 2D CNN utilising TL with signalencoded images and a 1D CNN model to improve the accuracy of mental wellbeing classification from the EnvBodySens dataset. This approach aims to overcome the challenges with data scarcity going beyond previous work by incorporating TL with signal encoded images in addition to a 1D CNN to classify additional sensor data.

TL first requires a pre-trained model; for this work we have explored multiple pre-trained networks. One model explored is MobileNet that uses a depthwise separable convolution, which is a depthwise convolution followed by a pointwise convolution, to reduce the model size and complexity, with a width multiplier and resolution multiplier used to tune the model. Additionally, batch normalisation is used instead of factorising convolutions. MobileNet has previously demonstrated competitive performance when compared to competing approaches in ILSVRC 2015 [55] using the Stanford dogs dataset [56] achieving 83.3% accuracy. Given that the majority of pre-trained models for TL have been trained on images it is beneficial to train these networks using signal encoded images from the continually changing motion data and not the physiological data which can often remain static resulting in little data being encoded.

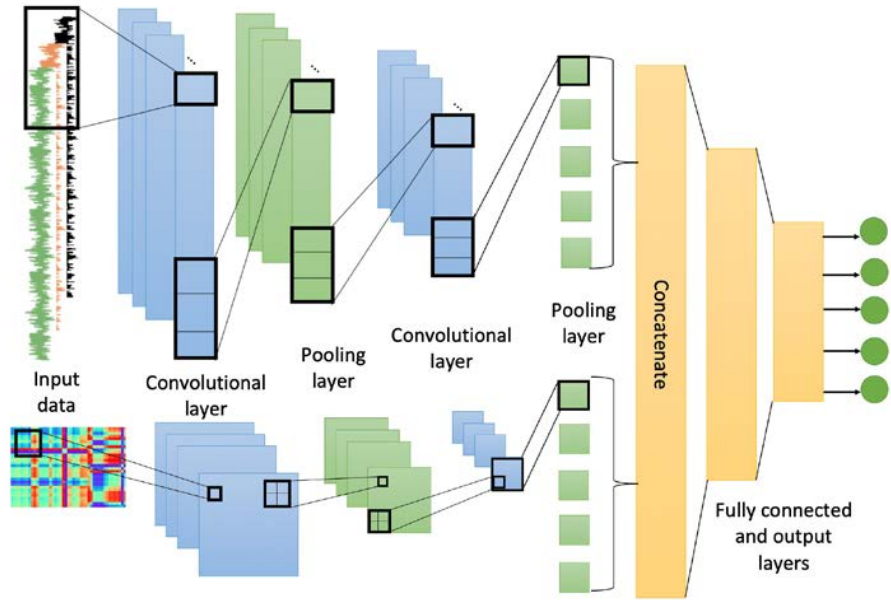

Figure 5: Combinatory model consisting of 1D CNN trained using raw physiological sensor data (top) and a 2D CNN using a transfer learning approach trained using accelerometer encoded images (bottom).

The approach transforms the accelerometer data in the EnvBodySens dataset into images using GADF, GASF and MTF, resulting in a total of 17,750 images for each encoding technique. These images are then used to train a 2D CNN classifying the 5 states of wellbeing exploring 7 pretrained models to apply the TL approach. An additional 1D CNN model using the same parameters as the aforementioned 1D CNN is trained using the remaining sensor data from the EnvBodySens dataset (HR, EDA, body temperature, acceleration, noise, UV and air pressure) to classify the 5 wellbeing states. The model is trained over 10 epochs with a batch size of 128. Batch normalisation and dropout layers have also been utilised to improve model accuracy and reduce overfitting. These two models are then combined as shown in figure 5, to explore the impact this signal-image encoding TL approach has on model performance when classifying wellbeing.

Hold-validation using a 20% test split has been used to test the model and calculate the F1-Score using around 284,000 sensor data frames for training and 71,000 for testing. Additionally, Leave-One-participant-Out CrossValidation (LOOCV) has also been utilised to test the TL approach on a subject-independent basis. The model is trained with 19 users' data then tested on the remaining user's data (19926 average data frames) to better simulate how the model would be used in the real-world to infer an individual's wellbeing.

## 5. Results

### 5.1. 1D CNN TL Stress Model - WISDM & Controlled Stressor data

As the controlled stressor experiment only collected physiological data (HR, HRV, EDA) it is not possible to use the same modality transformation techniques to transform fast changing motion data as used with the EnvBodySens dataset. However, it is possible to perform TL by initially training a similar model and then use a TL approach to re-train the last layers with the target stressor dataset.

Initially a 1D CNN model was trained using the HR, HRV and EDA data from the controlled experiment dataset, achieving a baseline accuracy of 82.5%. To evaluate whether TL can improve model performance the same 1D CNN was trained using the larger WISDM dataset consisting of accelerometer data for six human activities, achieving 91% accuracy. When training the physiological model using TL, the base WISDM model was imported without its last layer with dense layers then added to enable the new model to learn more complex functions. The dense layers used the relu activation function and the last layer, which contains two neurons, one for each class (stressed and relaxed) used the softmax activation function.

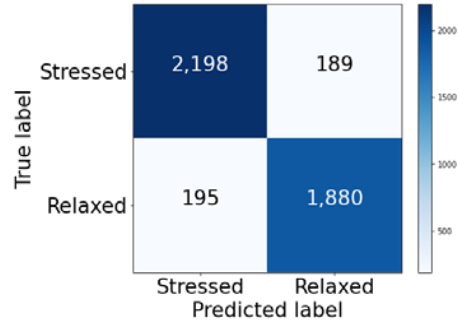

Figure 6: Confusion matrix of 1D CNN using TL to infer stressed and relaxed states of wellbeing from the controlled stressor dataset.

The model was then used to apply the TL approach. Once the model had been re-trained with the target controlled stressor dataset an accuracy of 91% was achieved using a 20% test split, an 8.5% accuracy improvement over the standard 1D CNN approach as shown in figure 6. This demonstrates the potential of TL to have a significant impact in improving the classification accuracy of mental wellbeing without the need to additionally collect extremely large datasets.

*5.2. Image Encoding Transfer Learning Model - EnvBodySens*

The second approach used the EnvBodySens dataset to explore the multiclass problem of classifying five emotional states using the signal-image encoding TL model. A range of 7 pre-trained models were used to explore the TL approach in addition to the 3 methods of signal-image transformation (GADF, GASF and MTF). Each of the approaches used transformed the motion data from the EnvBodySens dataset to images using a TL approach paired with a 1D CNN to train the remaining time-series data from the EnvBodySens dataset. The final testing F1-Scores on the 20% test data split are reported for each model in Table 1.

Table 1: Comparison of F1-Scores for different pre-trained DL models adapted for mental wellbeing inference through TL.

Table 1: Comparison of F1-Scores for different pre-trained DL models adapted for mental wellbeing inference through TL.

|  | GASF | | GADF | | MTF | |
|---|---|---|---|---|---|---|
|  | Physiological | All | Physiological | All | Physiological | All |
| Xception | 0.975 | 0.960 | 0.977 | 0.971 | 0.972 | 0.956 |
| VGG19 | **0.984** | 0.952 | 0.980 | 0.940 | 0.964 | 0.950 |
| ResNet | 0.963 | 0.955 | **0.978** | 0.937 | 0.964 | 0.972 |
| NasNet | 0.977 | 0.965 | **0.983** | 0.963 | 0.967 | 0.964 |
| DensetNet | 0.975 | 0.977 | **0.985** | 0.971 | 0.970 | 0.977 |
| MobileNetV2 | **0.981** | 0.970 | **0.981** | 0.954 | 0.979 | 0.970 |
| MobileNet | **0.980** | 0.967 | 0.968 | 0.955 | 0.974 | 0.959 |

*5.2.1. Data Modalities*

The data modalities were investigated to explore which modalities most contribute towards the classification of mental wellbeing. When all sensor data (HR, EDA, UV, body temperature, air pressure and noise) was used to train the 1D CNN combined with the TL approach for the signal-image transformed motion data, the model classified the 5 emotional stated between 93.7% and 97.7% accuracy. The 1D CNN was also trained using only physiological data (HR and EDA) to examine the impact not including environmental data has on model performance. When using the TL approach for motion data and a 1D CNN trained using only physiological data, the model accuracy increased to the highest achieved accuracy of 98.5% when using GADF to transform the motion data and DenseNet to perform TL, as shown in figure 7. Furthermore, when comparing the highest accuracy for each pre-trained CNN the physiological model consistently outperformed the model trained using all modalities. This demonstrates the importance of HR and EDA data when classing wellbeing unlike environmental data which resulted in more misclassification errors, in particular class 5 the poorest mental wellbeing state.

To evaluate whether the signal-image encoding TL approach improves model performance all sensor data (HR, EDA, noise, UV, body temperature, air pressure and accelerometer data) was used to train the 1D CNN model without performing TL or signal-image encoding. The 1D CNN achieved 93% accuracy, an overall reduction in accuracy compared with the TL model.

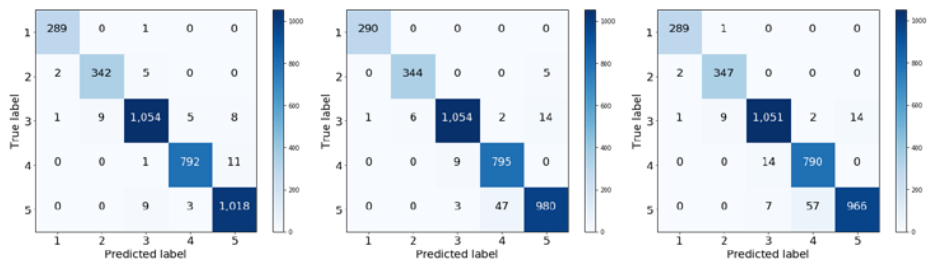

Figure 7: Confusion matrix for DenseNet model trained using HR, EDA and GADF encoded motion data (top left), GASF (top right) and MTF (bottom).

Furthermore, when the same 1D CNN was trained again using only HR and EDA data, the model achieved 94% accuracy, as shown in figure 8, a 4.5% reduction in accuracy. This demonstrates that image encoding and TL can increase overall model accuracy but performance is highly dependent on the addition sensor modalities used to train the network.

*5.2.2. Pre-trained CNN*

To explore whether the high accuracy achieved was influenced by the pretrained model used in the TL approach, other pre-trained CNNs were tested using the same GASF, GADF and MTF transformed images. As shown in table 1 DensNet achieved the highest accuracy for the GADF transformed data although VGG19 achieved the highest accuracy for GASF data and MobileNetV2 for MTF data. This demonstrates that the pre-tained model selected has little impact on performance with the average variance between the best and worst performing model for all 3 image encoding techniques being only 1.77% for the physiological models and 2.87% for the models trained using all sensor data. However, if the highest performance is required, it is necessary to test a range on pre-trained models on the individual dataset to ensure the most suitable model is selected.

*5.2.3. Signal-Image Encoding Technique*

The signal-image encoding technique also impacted model performance. GASF and GADF outperformed MTF for each pre-trained model, where GADF achieved the highest performance for four of the pre-trained models and GASF for the remaining three. The average accuracy for the GADF physiological model was 97.9% compared with 97.6% for GASF and 97% for MTF showing negotiable variations in performance between the different techniques.

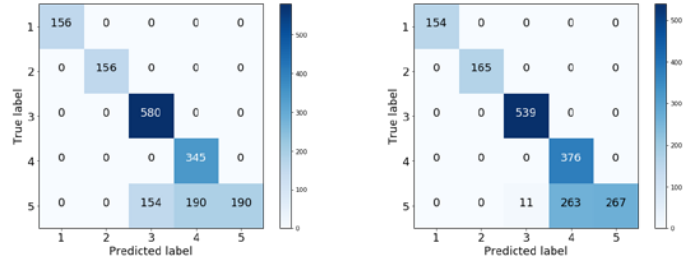

Figure 8: Confusion matrix of CNN model without TL when trained using all data achieving 93% accuracy (left) and only physiological and motion data achieving 94% accuracy (right).

*5.2.4. Subject-Independent Models*

As the GADF signal-encoding technique slightly outperformed the other encoders, it was used to explore subject-independent physiological models. Table 2 shows the accuracy achieved for each of the 20 users when the model was tested using LOOCV with each individual's physiological data. The accuracies range between 36.4% for user 1 and 77.7% for users 16 and 17. The outlier low accuracy for user 1 is due to corrupt EDA data which continually recorded null readings. The remaining users demonstrate more consistent accuracies and while lower than when tested using hold-out validation, they demonstrate the possibility of inferring wellbeing on an individual basis.

The subject-independent models were also trained without the TL approach while still transforming signals into images to explore whether performance improvements were due to the TL approach. A 2D CNN was implemented to train the signal encoded images which was concatenated with the 1D CNN trained using the physiological data. The results show TL only improved accuracy by an average of 0.55% for all users demonstrating little performance improvement. However, the TL approach never degraded model performance and achieved up to a 4% increase in accuracy showing it can be beneficial and should continue to be used.

Table 2: Comparison of subject-independent F1-Scores.

Table 2: Comparison of subject-independent F1-Scores.

| User | F1-score | User | F1-Score |
|---|---|---|---|
| 1 | 0.364 | 11 | 0.709 |
| 2 | 0.698 | 12 | 0.734 |
| 3 | 0.702 | 13 | 0.723 |
| 4 | 0.683 | 14 | 0.738 |
| 5 | 0.666 | 15 | 0.749 |
| 6 | 0.752 | 16 | 0.777 |
| 7 | 0.736 | 17 | 0.777 |
| 8 | 0.706 | 18 | 0.753 |
| 9 | 0.737 | 19 | 0.690 |
| 10 | 0.636 | 20 | 0.763 |

Overall, the combinatory approach of encoding accelerometer data as images, using a TL approach and pairing this with a 1D CNN trained using physiological data has improved the accuracy at which mental wellbeing can be classified. This approach successfully outperformed previous mental wellbeing classifiers achieving 98.5% accuracy when GADF was used to encode the accelerometer data, a pre-trained DenseNet was used to perform TL and phsyiological data was used to train the 1D CNN.

**6. Discussion**

We have introduced a new CNN-TL-based approach towards mental wellbeing classification going beyond previous signal-image encoding frameworks by incorporating TL in addition to a 1D CNN. We have demonstrated that a signal-image encoding TL approach can improve the performance in which 5 mental wellbeing states can be inferred, achieving up to 98.5% accuracy using hold-validation and an average of 72.3% using LOOCV. Furthermore, a TL approach to alleviate limited wellbeing data by transferring knowledge from a model trained using a large physical activity dataset improved accuracy to 91% when classifying stress.

The results have demonstrated that the integration of TL as part of the new proposed methodology extending standard DL algorithms can greatly improve the classification of mental wellbeing. In particular, encoding accelerometer data as images using GADF, GASF and MTF then using a pretrained model to perform TL and combining this model with a 1D CNN trained using physiological data, it was possible to improve performance by 4.5%. The encoding technique used was shown to have a minor impact in model accuracy demonstrating GADF was most effective for the majority of the pre-trained models. Similarly, the pre-trained model used to perform TL had a limited impact on model performance with an average difference of only 2.32% between the different models. However, TL only slightly improved performance by an average of 0.55% when testing using subject-independent signal-image encoded models, demonstrating the transformed images had a greater impact on model performance than TL.

Furthermore, we conclude that solely using physiological and motion data resulted in the highest accuracy (98.5%), outperforming models additionally trained using environmental data. This suggests that environmental factors such as noise and UV are more challenging to use for mental wellbeing inference even when paired with physiological data. The reduced performance may be due to the intricate information in the environmental data already being captured inherently in the physiological and motion data for example poor weather having a negative impact on mood.

When testing using LOOCV we found that the subject-independent accuracies are lower than subject-dependent accuracies. The average accuracy of the subject-independent physiological models excluding user 1 was 72.3% (SD 0.038), compared with 98.5% for the subject-dependent model both using GADF to transform the signals and a DenseNet pre-trained model. This likely reflects that different individuals have different patterns of physiology when experiencing the same emotions and that similar levels of activity are perceived differently in terms of valence [57] demonstrating similar results as other studies [58], [59], [60].

Furthermore, a second TL approach involved training a 1D CNN using a human activity dataset and then performing TL to re-train the model using physiological data from a controlled stressor experiment. This TL approach improved model accuracy by 8.5% over the base model, again demonstrating the benefits of TL approaches. However, this model did not perform as well as the image encoded TL model achieving 91% accuracy. This was unexpected as the EnvBodySens dataset used to train the model contains 5 classes and was conducted in the real-world where it is traditionally more challenging to classify wellbeing compared with the controlled experiment dataset which only contained 2 classes. This demonstrates the vital role motion data encoded as images plays in improving mental wellbeing classification accuracy, although it may not always be possible to collect motion data such as during the controlled stressor experiment where participants were stationary.

The achievement of 98.5% classification accuracy is highly encouraging for affective modelling, however there are a number of limitations and future directions that should be considered. While the EnvBodySens dataset includes physiological sensor data for emotion modelling and is representative of a typical affective modelling scenario, the TL approach needs to be tested on diverse datasets with more participants in various real-world scenarios to ensure its performance is sustained. Also, when using the controlled experiment dataset it would also be worth evaluating selective TL [61] where only the participants data that is similar to the target model is used to train the base model. Selective TL may help develop more personalised wellbeing models and further improve model performance. Furthermore, the presented framework could in principle be applied to explore additional modalities towards further improving the automatic estimation of mental health status in applications beyond emotion recognition.

Overall, three multivariate datasets were used as benchmarks datasets to evaluate the TL approaches. We have demonstrated using the proposed TL approaches that we can capitalize on two modalities to accurately classify wellbeing on a 5-point Likert scale. Our results have demonstrated that TL approaches are appropriate for modeling affective states especially when training data is scarce. These TL approaches have outperformed previous performances of models using the same EnvBodySens dataset built on ad-hoc extracted features [28] and 2 dimensional CNNs [27]. These findings showcase the potential for TL to improve affective multi-modal modeling with limited datasets.

## 7. Conclusion and Future Work

Recent developments in wearables and edge computing are producing sensory datasets as people are going about their daily activities. However, accurately classifying these datasets can be a challenging proposition especially when classifying mental wellbeing from physiological sensor data. In this study, we presented a scenario of wellbeing classification using small multimodal datasets. Although these types of time series datasets can help us understand people's emotions, current emotion recognition techniques are not efficient enough to tackle data scantiness. TL offers an automated way to utilise the learning outcomes from larger datasets. This paper has demonstrated the advantages of employing a combinatory TL approach for raw multimodal sensor data modelling.

Based on experimental results, the proposed frameworks employ 2 TL models using 3 multi-modal sensor datasets. The first TL approach trained a 1D CNN model using the WISDM dataset and performed a TL approach to adapt the model for the target stress dataset of physiological signals. This model achieved 91% accuracy, a 8.5% improvement over the same 1D CNN trained without TL. The second model used our proposed framework combining a TL approach using signal encoded images with a 1D CNN. Accelerometer data was transformed into RGB images using GADF, MTF and GASF. This data was subsequently used to train a 2D CNN using pretrained models to apply a TL approach. This model was concatenated with a 1D CNN architecture trained using raw physiological sensor data and resulted in increased performance, achieving 98.5% accuracy. Incorporating a TL signal-image encoding approach in addition to a 1D CNN helped further improve model performance with limited data.

There are several future directions to further study the TL approaches used in this research. First, this work has only explored 1D CNNs paired with the TL model, in the future it would be worth evaluating whether other architectures could further improve classification accuracy. Additionally, this framework could be explored beyond the classification of mental wellbeing using alternative datasets to evaluate performance in other domains.

[39] B. Shickel, M. Heesacker, S. Benton, P. Rashidi, Hashtag Healthcare: From Tweets to Mental Health Journals Using Deep Transfer Learning (aug 2017).

[40] F. Fahimi, Z. Zhang, W. B. Goh, T. S. Lee, K. K. Ang, C. Guan, Intersubject transfer learning with an end-to-end deep convolutional neural network for EEG-based BCI, Journal of Neural Engineering (2019). doi: 10.1088/1741-2552/aaf3f6.

[41] H. W. Ng, V. D. Nguyen, V. Vonikakis, S. Winkler, Deep learning for emotion recognition on small datasets using transfer learning, in: ICMI 2015 - Proceedings of the 2015 ACM International Conference on Multimodal Interaction, Association for Computing Machinery, Inc, New York, New York, USA, 2015, pp. 443–449. doi:10.1145/2818346. 2830593.

[42] X. Ouyang, S. Kawaai, E. G. H. Goh, S. Shen, W. Ding, H. Ming, D. Y. Huang, Audio-visual emotion recognition using deep transfer learning and multiple temporal models, in: ICMI 2017 - Proceedings of the 19th ACM International Conference on Multimodal Interaction, Vol. 2017-

Janua, Association for Computing Machinery, Inc, New York, New York, USA, 2017, pp. 577–582. doi:10.1145/3136755.3143012.

[43] K. Dedovic, R. Renwick, N. K. Mahani, V. Engert, The Montreal Imaging Stress Task : using functional imaging to investigate the ..., Psychiatry & Neuroscience 30 (5) (2005) 319–325.

[44] S. G. Hart, L. E. Staveland, Development of NASA-TLX (Task Load Index): Results of Empirical and Theoretical Research, Advances in Psychology (1988). doi:10.1016/S0166-4115(08)62386-9.

[45] F. R. H. Zijlstra, L. Doorn, The construction of a scale to measure subjective effort, Delft, The Netherlands: Delft University of Technology, Department of Philosophy and Social Sciences (1985).

[46] M. M. Bradley, P. J. Lang, Measuring emotion: The self-assessment manikin and the semantic differential, Journal of Behavior Therapy and Experimental Psychiatry (1994). doi:10.1016/0005-7916(94) 90063-9.

[47] D. Gould, Visual Analogue Scale (VAS) - Information Point, Journal of Clinical Nursing (2001). doi:10.4193/Rhino10.303.

[48] E. Banzhaf, F. De La Barrera, A. Kindler, S. Reyes-Paecke, U. Schlink,J. Welz, S. Kabisch, A conceptual framework for integrated analysis of environmental quality and quality of life, Ecological Indicators (2014). doi:10.1016/j.ecolind.2014.06.002.

[49] Microsoft, Microsoft Band 2 features and functions (2019). URL support.microsoft.com/en-gb/help/4000313
28